\documentclass[twoside]{article}

\usepackage[T1]{fontenc}

\usepackage[utf8]{inputenc} 
\usepackage{tgtermes}
\usepackage{amssymb}
\usepackage{amsfonts}

\usepackage{tgtermes}
\usepackage{amsmath}
\usepackage{scalefnt,letltxmacro}
\LetLtxMacro{\oldtextsc}{\textsc}
\renewcommand{\textsc}[1]{\oldtextsc{\scalefont{1.10}#1}}
\usepackage[scaled=0.92]{PTSans}

\usepackage[usenames,dvipsnames]{xcolor}
\definecolor{shadecolor}{gray}{0.9}

\usepackage[expansion=alltext]{microtype}
\usepackage[english]{babel}
\usepackage[parfill]{parskip}
\usepackage{afterpage}
\usepackage{framed}
\usepackage{nicefrac}
\usepackage{bm}
\usepackage{fixmath}

\usepackage[colorinlistoftodos,
           prependcaption,
           textsize=small,
           backgroundcolor=yellow,
           linecolor=lightgray,
           bordercolor=lightgray]{todonotes}

\usepackage{lineno}

\usepackage{ragged2e}

\usepackage{graphicx}
\usepackage[labelfont=bf]{caption}
\usepackage[format=hang]{subcaption}

\usepackage{booktabs}
\usepackage{arydshln} 
\usepackage{natbib}

\usepackage{listings}
\usepackage{fancyvrb}
\fvset{fontsize=\normalsize}

\usepackage[colorlinks,linktoc=all]{hyperref}
\usepackage[all]{hypcap}
\hypersetup{citecolor=Violet}
\hypersetup{linkcolor=black}
\hypersetup{urlcolor=MidnightBlue}
\usepackage{url}

\usepackage[acronym,smallcaps,nowarn]{glossaries}

\usepackage{listings}
\lstdefinestyle{alp_style}{
    commentstyle=\color{OliveGreen},
    numberstyle=\tiny\color{black!60},
    stringstyle=\color{BrickRed},
    basicstyle=\ttfamily\scriptsize,
    breakatwhitespace=false,
    breaklines=true,
    captionpos=b,
    keepspaces=true,
    numbers=none,
    numbersep=5pt,
    showspaces=false,
    showstringspaces=false,
    showtabs=false,
    tabsize=2
}
\usepackage{bbold}

\usepackage{mathtools}

\DeclareMathOperator*{\argmax}{arg\,max}

\def\independenT#1#2{\mathrel{\rlap{$#1#2$}\mkern2mu{#1#2}}}

\newcommand{\mba}{\mathbold{a}}
\newcommand{\mbb}{\mathbold{b}}

\newcommand{\mbx}{\mathbold{x}}

\newcommand{\mbz}{\mathbold{z}}

\newcommand{\mbbeta}{\mathbold{\beta}}

\newcommand{\mblambda}{\mathbold{\lambda}}

\newcommand{\mbphi}{\mathbold{\phi}}
\newcommand{\mbpi}{\mathbold{\pi}}

\newcommand{\mbtheta}{\mathbold{\theta}}

\newcommand{\mbvarphi}{\mathbold{\varphi}}

\newcommand{\mbOmega}{\boldsymbol{\mathrm{\Omega}}}

\newacronym{BBVI}{bbvi}{black box variational inference}
\newacronym{ELBO}{elbo}{evidence lower bound}
\newacronym{VI}{vi}{variational inference}
\newacronym{KL}{kl}{Kullback-Leibler}
\newacronym{MCMC}{mcmc}{Markov chain Monte Carlo}
\newacronym{klVI}{klvi}{$\operatorname{KL}(q || p)$ variational inference}
\newacronym{VAE}{vae}{variational autoencoder}
\newacronym{SAVAE}{sa-vae}{semi-amortized variational autoencoder}
\newacronym{skipVAE}{skip-vae}{Skip Variational Autoencoder}
\newacronym{SkipSAVAE}{skip-sa-vae}{Skip Variational Autoencoder}
\newacronym{LM}{language model}{Language Model}
\newacronym{GSM}{gsm}{generative skip model}
\newacronym{MLP}{mlp}{multilayer perceptron} 
\newacronym{GumbelSoftmaxVAE}{gsm-vae}{gumbel softmax VAE} 
\newacronym{BernBPE}{bern-bpe}{Bernoulli likelihood Beta process ecnoder} 
\newacronym{GaussBPE}{gauss-bpe}{Poisson likelihood Beta process ecnoder} 

\usepackage{bbm}

\usepackage{multirow} 
\usepackage[accepted]{aistats2021}

\hypersetup{citecolor=Violet}
\hypersetup{linkcolor=black}
\hypersetup{urlcolor=MidnightBlue}

\newcommand{\ts}{\textsuperscript}
\usepackage{cleveref}
\newcommand\independent{\protect\mathpalette{\protect\independenT}{\perp}}
\def\independenT#1#2{\mathrel{\rlap{$#1#2$}\mkern2mu{#1#2}}}
\usepackage{dblfloatfix}
\usepackage{upgreek}
\newcommand\numberthis{\addtocounter{equation}{1}\tag{\theequation}}

\usepackage{algorithm}
\usepackage[noend]{algpseudocode}
\usepackage{etoolbox}
\usepackage{tikz}
\usetikzlibrary{tikzmark}
\usetikzlibrary{calc}

\begin{document}

%

%

\twocolumn[

\aistatstitle{Bayesian Beta-Bernoulli Process Sparse Coding with Deep Neural Networks}

\aistatsauthor{Arunesh Mittal \And Kai Yang \And Paul Sajda \And  John Paisley}
 
\aistatsaddress{ Columbia University  \And  Columbia University \And Columbia University \And Columbia University } ]

\begin{abstract}
  Several approximate inference methods have been proposed for deep discrete latent variable models. However, non-parametric methods which have previously been successfully employed for classical sparse coding models have largely been unexplored in the context of deep models. We propose a non-parametric iterative algorithm for learning discrete latent representations in such deep models. Additionally, to learn scale invariant discrete features, we propose local data scaling variables. Lastly, to encourage sparsity in our representations, we propose a Beta-Bernoulli process prior on the latent factors. We evaluate our spare coding model coupled with different likelihood models. We evaluate our method across datasets with varying characteristics and compare our results to current amortized approximate inference methods.
\end{abstract}

\section{Introduction}
Sparse coding \citep{olshausen1996emergence} is an unsupervised latent factor model that has been widely used to uncover sparse discrete latent structure from data. Unlike auto-encoders, where the encoder is a parametric model, the encoder in sparse coding is an optimization algorithm that searches for an optimal encoding $\mbz^* = \argmax_\mbz p(\mbx ,\mbz; \theta)$, which maximizes the joint likelihood of the data  $\mbx$ and latent encodings $\mbz$. An advantage of the non-parametric approach is that it decouples the encoder and decoder such that the generalization error rises entirely from the reconstruction error of the decoder. In such models, sparsity is encouraged in the latent encodings  $\mbz$ via a  prior such as Laplace, Cauchy or factorized Student-t prior \citep{goodfellow2016deep}. Sparse coding by optimizing the MAP objective with a Laplace prior allows one to use gradient optimization methods for inferring $\mbz$. However, one major drawback using such priors is that the latent factors in the encoding are encouraged to remain close to zero, even when those factors are active, whereas, for inactive elements, under the prior distribution, a factor being exactly zero has zero probability \citep{goodfellow2012scaling}. 


Variational Auto Encoders \citep{kingma2013auto, MAL-056} have been  popular deep generative models employed to uncover lower dimensional latent structure in data. Despite the flexibility of the deep likelihood model $p(\mbx \mid \mbz)$, VAEs use a parametric encoder network for inferring the latent encoding $\mbz$, and hence do not benefit from the same advantages as that of a non-parametric encoding model. In VAEs, the generalization error is linked to both the decoder and the encoder and is difficult to disentangle. In addition to the limitations of using a parametric network for inference, amortized variational inference using parametric neural networks has additional learning constraints due to the amortization and approximation gaps in the variational objective used to train VAEs \citep{cremer2018inference}. In principle, a non-parametric encoding model with a MAP-EM optimization can perform better than neural net parameterized ammortized inference, as it does not suffer from the amortization gap or the variational approximation gap. This comes at the cost of losing posterior uncertainty estimates, however, this might be an acceptable trade-off given that the posterior uncertainty in deep generative models via ammortized approximate inference is poorly calibrated and is still an area of active research \citep{nalisnick2018deep}. Additionally, utilizing the  MAP estimates, we can still potentially approximate posterior uncertainty using a Laplace approximation \citep{ritter2018scalable}.

VAE models with discrete latent factors \citep{maddison2016concrete, jang2016categorical}, do not work well with continuous data likelihood models, as the discrete sparse latent factors have limited representational capacity, and are unable to adequately represent local scale variations across an entire dataset. In fact, often one desires that the latent encodings only encode underlying latent structure of the data that is invariant to local data point scale variations.

To address the aforementioned issues, we propose a generative model with local scaling variables that decouples the data scaling from the discrete latent representation. We utilize a Beta-Bernoulli process prior on the latent codes that allows us to learn sparse discrete latent factors. For inference in this model, we propose a MAP-EM greedy pursuit algorithm. We expect the inferred latent codes with true zeroes to have a stronger regularizing effect than the above mentioned sparsity promoting priors, which is especially advantageous in deep generative models with flexible neural network parameterized likelihood models. The primary disadvantage of the non-parametric encoder is that it requires greater time to compute $\mbz$ due to the iterative algorithm, however, since the Beta-Bernoulli prior encourages the encodings to be sparse, as training progresses, the time taken to encode each data point significantly decreases over training iterations. 

We demonstrate the efficacy of our model by proposing three different instantiations of our general model. We evaluate our models on discrete and continuous data by examining the representational capacity of our model by measuring the data reconstruction error, as well as the sparsity of our learned representations. We compare our models to widely used VAE \citep{kingma2013auto} and its discrete variant the Gumbel Softmax VAE \citep{jang2016categorical}. Not only does our model perform better in terms of reconstruction errors, it also learns substantially sparser latent encodings.


\section{Related Work}
We briefly review the VAE model that has been widely used to learn latent representations. In the typical VAE generative model, $\mbz_n$ is drawn from a Gaussian prior, then given $\mbz_n$, $\mbx_n$ is then drawn from a distribution parametrized by a deep neural network $f_\theta(\cdot)$, which maps $\mbz_n$ to the sufficient statistics of the likelihood function $p(\mbx_n \mid \mbz_n)$:
\begin{align*}
    \mbz_n &\sim p(\mbz_n)  \\
    \mbx_n &\sim p(\mbx_n \mid f_\theta(\mbz_n); \mbtheta)  
\end{align*}
Inference in this model is then performed using variational inference, however, unlike free form optimization used with mean field variational inference \citep{jordan1999introduction}, the VAE models, parametrize the variational $q(\mbz_n; \mbphi)$ distribution also with a neural network, that maps the data $\mbx_n$ to the sufficient statistics of the $q(\cdot)$ distribution. Then the posterior inference is performed by optimizing the Evidence Lower Bound ELBO using gradient methods:
\begin{align*}
    \ln p(\mbx) \geq \mathrm{ELBO} = \sum_n \mathbb{E}_{q(\mbz_n | \mbx_n ; \mbphi)}\left[\ln \frac{p(\mbx_n , \mbz_n; \mbtheta)}{q(\mbz_n \vert \mbx_n; \mbphi)}\right]
\end{align*}

\section{Beta-Bernoulli Generative Process}
We propose the following generative model with Beta-Bernoulli process prior. Given observed data $\mbx_n$, the corresponding latent encoding $ \mbz_n$ is drawn from a Bernoulli process (BeP) parameterized by a beta process (BP), where, the Bernoulli process prior over each of the $k$ factors $\mbz_{nk} \in \mbz_n$, is parameterized by $\mbpi_k$ drawn from a finite limit approximation to the beta process \citep{griffiths2011indian, paisley2009nonparametric}. Since $\mbz_{nk}$ is drawn from $\mathrm{Bern}(\mbpi_k)$, where $\mbpi_k \sim \mathrm{Beta}\left(\alpha \gamma / K, \alpha \left( 1- \gamma\ K \right)\right)$, $k \in \{1,\ldots,K\}$, the random measure $G_{n}^{K} = \sum_{k=1}^{K} \mbz_{nk} \delta_{f_\theta(\mbz_{nk})}$, $\lim_{K\to\infty} G_{n}^{K}$ converges to a Bernoulli process \citep{paisley2016constructive}.

Then given a latent binary vector $\mbz_n \in \{0,1\}^K$, the observed data point $\mbx_n$ is drawn from an exponential family distribution with a local scaling factor $\mblambda_n$, also drawn from an appropriate exponential family distribution (\crefrange{section:BernBEP}{section:GaussBEP}). The natural parameters of this data distribution are parametrized by a $L$ layered neural network $f_\theta(\cdot)$. The neural network $f_\theta(\cdot)$, maps the binary latent code $\mbz_n \in \{0,1\}^K$ to $\mathbb{R}^D$. This corresponds to the following generative process: 
\begin{align*}
    \mbpi_k  &\sim \mathrm{Beta}\left(\alpha (\gamma / K), \alpha \left( 1- (\gamma / K)\right)\right) \\
    \mbz_{nk} &\sim \mathrm{Bern}(\mbpi_k)\\  
    \mblambda_n  &\sim  \mathrm{ExpFam}(\mbphi) \\
    \mbx_n &\sim  \mathrm{ExpFam}(f_\theta(\mbz_n); \mblambda_n)
\end{align*}
where, $\mbpi_k$ is the global prior on $\mbz_{nk}$, which corresponds to the $k^{th}$  dimension of the latent encoded vector $\mbz_{n}$, and $\mbz_{n}$ is the local latent encoding for the $n^{th}$ data point $\mbx_n$. The likelihood model for $\mbx$ is parametrized by local parameters $\{\mblambda_n\}_{n=1}^{N}$ and the global parameters $\mbtheta$.

During inference, the Beta-Bernoulli process prior on $\mbz$, encourages the model to learn sparse latent encodings. As we would like the binary encodings to be scale invariant, modeling local data point specific scale distibution $p(\mblambda_n)$, allows us to marginalize out the scale variations in data when when inferring the latent code $\mbz_n$. We demonstrate the utility of this non-parametric encoding model by coupling the Beta-Bernoulli process sparse encoding prior with three distinct exponential family likelihood models in the following sections.

\subsection{Scale Invariant Models}
Given two data points $\mbx_{m}$ and $\mbx_{n}$, where $\mbx_{m}$ is just a scaled version of $\mbx_{n}$, we would want these data points to have the same latent embedding $\mbz$. To disentangle the scale of the data points from the latent discrete representation, we introduce a local scale distribution for the Gaussian and Poisson likelihood models.
\subsubsection{GaussBPE}
\label{section:GaussBEP}
For real valued data, we use a Gaussian likelihood model, where $f_\theta(\cdot)$ parametrizes the mean of the Gaussian distribution. We model the local data point scale $\mblambda_n$ with a univariate Gaussian:
\begin{align*}
    \mblambda_n &\sim \mathcal{N}(0, c) \\ 
    \mbx_n &\sim  \mathcal{N}(\mblambda_ n f_\theta(\mbz_n) , \sigma^2I)
\end{align*}

\begin{figure*}[ht]
    \centering
    \includegraphics[width=.95\textwidth, trim={4.cm 8cm 4cm 8cm},clip ]{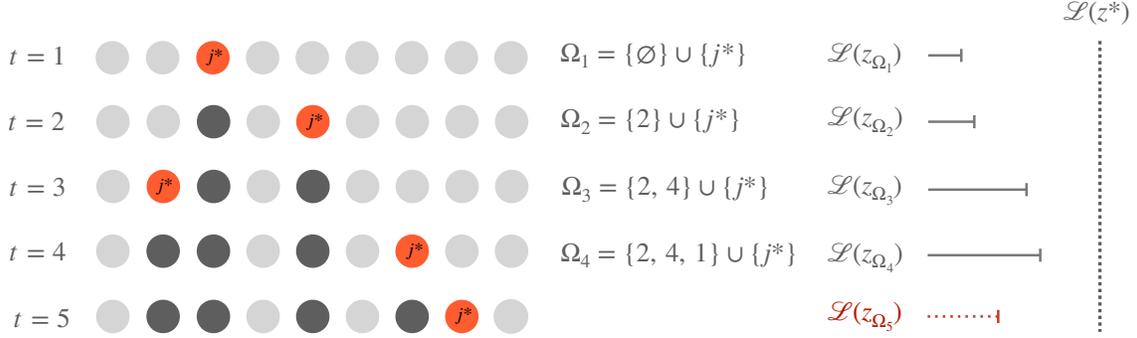}
    \caption{Greedy pursuit for $\mbz$. For each greedy sparse coding step, first all bits in the active set $\mbOmega$ are turned on, then individually all $j \in \{0, \ldots, K \} \setminus \mbOmega  $ are turned on. The bit $j^*$ that leads to maximal increase in the bound $\mathcal{L}(\mbz_{\mbOmega \cup \{j^*\}})$ is added to the active set $\mbOmega$. If adding a $j^*$ leads to a decrease in the bound, the search is terminated and the sparse vector $\mbz_{\mbOmega}$ is returned as the sparse code. Here $\mathcal{L}(\mbz^*)$ is the optimal encoding that could be recovered if an exhaustive search were performed over $2^K$ possible codes. }
    \label{fig:illus}
\end{figure*} 

The likelihood given the local encoding $\mbz_n$, depends on both local parameters $\mblambda_n$ and the global neural net parameters $\mbtheta$. For $\mblambda = 1$, this is equivalent to the isotropic Gaussian likelihood with Gaussian prior generative model, employed by \citet{kingma2013auto}.

\subsubsection{PoissBPE}
\label{section:PoissBEP}
For count data we use a Poisson likelihood model, where $f_\theta(\cdot)$ parametrizes the rate of the Poisson distribution. We model the local data point rate $\mblambda_n$ with a Gamma distribution. Additionally, we introduce a global parameter $\mbbeta$:
\begin{align*}
    \mblambda_n &\sim \mathrm{Gamma}(a, b) \\ 
    \mbx_n &\sim  \mathrm{Poiss}(\mblambda_ n \mbbeta f_\theta(\mbz_n))  
\end{align*}
The likelihood then depends on local parameters $\{\mblambda_n\}_{n=1}^{N}$ and the global neural net parameters $\mbvarphi$, and $\mbbeta$, a $W \times T$ matrix, where each column $\mbbeta_{:,t} \in \Delta_{W-1}$. The global parameters include both the neural net parameters $\mbvarphi$ and $\mbbeta$, $\mbtheta = \{\mbvarphi, \mbbeta\}$

In the context of topic modeling, $W$ corresponds to number of words in the vocabulary, $T$ corresponds to the number of topics, and $\mbbeta_{:,t}$, corresponds to the $t^{th}$ topic distrubution over words. Then $\mbx_n^{(d)}$ is the number of occurances of word $d$ in the $n^{th}$ document.

\subsection{BernBPE}
\label{section:BernBEP}
To evaluate a likelihood model, where we do not need to explicitly model the local scale, such a binary data, we use a Bernoulli likelihood model, where $f_\theta(\cdot)$ parametrizes the mean of the Bernoulli distribution, without any local scaling variables:
\begin{align*}
    \mbx_n &\sim  \mathrm{Bern}(f_\theta(\mbz_n)) 
\end{align*}
Given the local encoding $\mbz_n$, the likelihood model only depends on the global neural net parameters $\mbtheta$. This model is equivalent to the Bernoulli likelihood model with Bernoulli prior employed by \citet{jang2016categorical}.

\section{Inference}
We propose a MAP-EM algorithm to perform inference in this model. We compute point estimates for local latent encodings $\{\mbz_n\}_{n=1}^{N}$ and the global parameters $\mbtheta$ and compute posterior distributions over $\mbpi$ and $\mblambda$. Since, $\mbpi \independent \mbx_n \mid \mbz_n$, utilizing the conjugacy in the model, we can analytically compute the conditional posterior $q(\mbpi)$. Similarly, for the local scaling variables in the Gaussian likelihood and Poisson likelihood models, we can analytically compute the conditional posterior $q(\mblambda_n)$. 

\subsection{Inference for local scale parameters}
\subsubsection{$q(\mblambda)$ for Gaussian Likelihood}
Since the conditional posterior $q(\mblambda) \triangleq p(\mblambda \mid \mbx, \mbtheta, \mbz)$ factorizes as $\prod_n q(\mblambda_n) = \prod_n  p(\mblambda_n \mid \mbx_n, \mbtheta, \mbz_n)$ and the posterior distribution over $\mblambda_n$ is also a Gaussian, we can analytically compute the posterior $q(\mblambda_n)$:
\begin{align*}
        q(\mblambda_n) &= 
            \mathcal{N}(\mblambda_n \mid \mu_{\mblambda_n \mid \mbx_n,  \mbz_n, \mbtheta},            \sigma^{2}_{\mblambda_n \mid \mbx_n, \mbz_n, \mbtheta})  \numberthis \label{eqn:lmdba_gauss}\\
        \sigma^{2}_{\mblambda_n \mid \mbx_n, \mbz_n, \mbtheta} 
            &= \left( c^{-1} + f_\theta(\mbz_n)^{\top} f_\theta(\mbz_n) / \sigma^2 \right)^{-1} \\
        \mu_{\mblambda_n \mid \mbx_n, \mbz_n, \mbtheta}
            &= (\sigma^{2}_{\mblambda_n \mid \mbx_n, \mbz_n,\mbtheta})
               (f_\theta(\mbz_n)^{\top} \mbx_n)/ \sigma^2 
\end{align*}

\subsection{$q(\mblambda)$ for Poisson Likelihood}
The conditional posterior $q(\mblambda) \triangleq p(\mblambda \mid \mbx, \mbz, \mbtheta, \mbbeta)$ factorizes as $\prod_n q(\mblambda_n) = \prod_n  p(\mblambda_n \mid \mbx_n, \mbz_n, \mbtheta, \mbbeta)$. Given the Gamma prior on $\mblambda_n$, the posterior distribution over $\mblambda_n$ is also Gamma distributed, hence, we can analytically compute the posterior $q(\mblambda_n)$:
\begin{align*}
        q(\mblambda_n) &\propto  \mblambda_n^{a-1} e^{-b\mblambda_n}\textstyle\prod\nolimits_{w}  \numberthis \label{eqn:lmdba_poiss} (\mblambda_n)^{\mbx_n^{(w)}}e^{(-\mblambda_n \mbphi_n^{(w)})} \\
        &= \mathrm{Gamma}\left(\sum_w \mbx_n^{(w)} + a, b + 1\right)
\end{align*}
Where $\mbphi_n^{(d)} \triangleq [\mbbeta f_\theta(\mbz_n)]_d$. Since $\sum_t [f_\theta(\mbz_n)]_{t} \triangleq 1$ and $\sum_w \mbbeta_{w, t} \triangleq 1$, the sum over the random vector, $\sum_w \mbphi_n^{(w)} = 1$. Hence, the posterior $q(\mblambda_n)$ does not depend on $f_\theta(\mbz_n)$ or $\mbbeta$. In practice, we only need to compute $q(\mblambda)$ for the entire dataset just once during training.

\subsection{Inference for latent variables}
\subsubsection{Stochastic update for $q(\mbpi)$}
For scalable inference, given a batch of data $\{\mbx_n\}_{n \in \mathcal{B}}$, we first compute the latent codes $\{\mbz_n\}_{n \in \mathcal{B}}$, then we can efficiently compute the posterior $q(\mbpi \mid \{\mbz_n\}_{n \in \mathcal{B}})$ using natural gradient updates \citep{hoffman2013stochastic}. This posterior parameter update is a stochastic gradient step along the natural gradient, which is equivalent to following updates to the posterior sufficient statistics $\{\mba_k\}_{k=1}^{K}$ and  $\{\mbb_k\}_{k=1}^{K}$ with step size $\eta$:
\begin{align*}
        q(\mbpi) &= \textstyle\prod\nolimits_{k} \mathrm{Beta}(\mbpi_k \mid a_k, b_k)  \numberthis \label{eqn:pi} \\
        \mba_k^{\prime} &= \alpha \textstyle\frac{\gamma}{K} + \frac{N}{|S|}\sum\nolimits_{n \in S} \mbz^{(n)}_{k} \\
        \mbb_k^{\prime} &= \alpha \textstyle \left(1-(\gamma / K)\right) + \frac{N}{|S|} \sum_{n \in S} 
                       \left(1-\mbz^{(n)}_{k}\right) \\
        \mba_k &\gets (1-\eta) \mba_k + \eta \ \mba_k^{\prime}  \\
        \mbb_k &\gets (1-\eta) \mbb_k + \eta \ \mbb_k^{\prime}  \label{eq:q_pi_ab}
\end{align*}
\subsection{Greedy pursuit for $\mbz$}
For each model we marginalize the local scale variables $\mblambda$ and global $\mbpi$ to compute the complete data joint likelihood lower bounds $\mathcal{L}_\mathcal{G}, \mathcal{L}_\mathcal{P}, \mathcal{L}_\mathcal{B}$ for Gaussian, Poisson and Bernoulli likelihood models respectively, which include terms that only depend on $\mbz$:
\begin{align*}
    \mathcal{L_G}(\mbz) &= 
        \ln  \int p(\mbx,\mblambda \mid  \mbtheta, \mbz)  d\lambda + \mathbb{E}_{q(\mbpi)}\left[\ln 
        p(\mbz \mid \mbpi) \right] \\
    \mathcal{L_P}(\mbz) &= 
        \mathbb{E}_{q(\mblambda)}\left[\ln p(\mbx,\mblambda \mid  \mbtheta, \mbz) \right] + \mathbb{E}_{q(\mbpi)}\left[\ln 
        p(\mbz \mid \mbpi) \right] \\
    \mathcal{L_B}(\mbz) &= 
        \ln p(\mbx \mid \mbtheta, \mbz) + \mathbb{E}_{q(\mbpi)}\left[\ln 
        p(\mbz \mid \mbpi) \right]
\end{align*}

The expected log prior is given by:
\begin{align*}
\mathbb{E}_{q(\mbpi)}\left[\log  p(\mbz_n \mid \mbpi) \right] & = \textstyle\sum\nolimits_{k} \mbz_{nk} [\psi(\mba_k - \psi(\mba_k + \mbb_k)] \\
&+ (1-\mbz_{nk}) [\psi(\mbb_k) - \psi(\mba_k + \mbb_k)]
\end{align*}
where $\psi(\cdot)$ is the digamma function.

For the Gaussian likelihood model we can marginalize $\mblambda_n$ when maximizing $p(\mbx_n, \mblambda_n, \mbz_n)$:
\begin{align*}
    \mathcal{{L}_G}(\mbz_n) =  
         \ln \int p(\mbx_n, \mblambda_n \mid \mbtheta, \mbz_n) d\mblambda_n + \\ \mathbb{E}_{q(\mbpi)}\left[\ln p(\mbz_n \vert \mbpi) \right] 
\end{align*}
where the marginal log likelihood can be calculated:
\begin{align*}
    \ln &  \int p(\mbx_n, \mblambda_n \mid \mbtheta, \mbz_n)  d\mblambda_n = \\ 
        & -\frac{1}{2} \bigg[ \ln \bigg(1 + \tfrac{c}{\sigma^2} f_\theta(\mbz_n)^{\top} f_\theta(\mbz_n) \bigg)\\
        & \hspace{2em}+ \mbx_n^{\top} \left( \sigma^{-2}I - \tfrac{\frac{1}{\sigma^{2}} f_\theta(\mbz_n)f_{\theta}(\mbz_n)^{\top}}{c^{-1}\sigma^2 + f_{\theta}(\mbz_n)^{\top}f_{\mbtheta}(\mbz_n)}\right) \mbx_n \bigg]  
\end{align*}

\renewcommand{\thealgorithm}{}
\newcommand*{\Break}{\textbf{break}}
\algnewcommand{\And}{\textbf{and}}
\algnewcommand{\algorithmicgoto}{\textbf{go to}}%
\algnewcommand{\Goto}[1]{\algorithmicgoto~\ref{#1}}%

\errorcontextlines\maxdimen

\newcommand{\ALGtikzmarkcolor}{black}
\newcommand{\ALGtikzmarkextraindent}{4pt}
\newcommand{\ALGtikzmarkverticaloffsetstart}{-.5ex}
\newcommand{\ALGtikzmarkverticaloffsetend}{-.5ex}
\makeatletter
\newcounter{ALG@tikzmark@tempcnta}

\newcommand\ALG@tikzmark@start{%
    \global\let\ALG@tikzmark@last\ALG@tikzmark@starttext%
    \expandafter\edef\csname ALG@tikzmark@\theALG@nested\endcsname{\theALG@tikzmark@tempcnta}%
    \tikzmark{ALG@tikzmark@start@\csname ALG@tikzmark@\theALG@nested\endcsname}%
    \addtocounter{ALG@tikzmark@tempcnta}{1}%
}

\def\ALG@tikzmark@starttext{start}
\newcommand\ALG@tikzmark@end{%
    \ifx\ALG@tikzmark@last\ALG@tikzmark@starttext
    \else
        \tikzmark{ALG@tikzmark@end@\csname ALG@tikzmark@\theALG@nested\endcsname}%
        \tikz[overlay,remember picture] \draw[\ALGtikzmarkcolor] let \p{S}=($(pic cs:ALG@tikzmark@start@\csname ALG@tikzmark@\theALG@nested\endcsname)+(\ALGtikzmarkextraindent,\ALGtikzmarkverticaloffsetstart)$), \p{E}=($(pic cs:ALG@tikzmark@end@\csname ALG@tikzmark@\theALG@nested\endcsname)+(\ALGtikzmarkextraindent,\ALGtikzmarkverticaloffsetend)$) in (\x{S},\y{S})--(\x{S},\y{E});%
    \fi
    \gdef\ALG@tikzmark@last{end}%
}

\apptocmd{\ALG@beginblock}{\ALG@tikzmark@start}{}{\errmessage{failed to patch}}
\pretocmd{\ALG@endblock}{\ALG@tikzmark@end}{}{\errmessage{failed to patch}}
\makeatother

\makeatletter
\renewcommand{\ALG@name}{Algorithm 1:}
\makeatother

\makeatletter
\expandafter\patchcmd\csname\string\algorithmic\endcsname{\itemsep\z@}{\itemsep=.1ex}{}{} 
\makeatother

\makeatletter
\renewcommand{\ALG@beginalgorithmic}{\footnotesize}
\makeatother

\definecolor{dimgray}{rgb}{0.41, 0.41, 0.41}
\algnewcommand{\LeftComment}[1]{\State \(\triangleright\) #1}

\begin{algorithm}
\caption{Stochastic Pursuit Sparse Coding}
\begin{algorithmic}
    
     
    \While{\textbf{not} \text{converged}}
        \label{alg:sparse_code}   
        \State $\mathcal{B} \subset \{1, \ldots, N\}$   
        \\
        \LeftComment{\textcolor{dimgray}{Natural gradient updates for local parameters $\mblambda$}}
        \For{$n \in \mathcal{B}$}    
            \State Update $q(\mblambda_n)$ suff. stats 
        \EndFor
        \\
        \LeftComment{\textcolor{dimgray}{Natural gradient updates for global parameter $\mbpi$}}
        \For{$k \in \{1 \ldots K\}$}    
            \State Update $q(\mbpi_k)$ suff. stats 
        \EndFor
        \\
        \LeftComment{\textcolor{dimgray}{Greedy pursuit for $\mbz$ to maximize $\mathcal{L}_{(\cdot)}(\mbz)$}} 
        \State \textbf{Initialize:} $\forall \, (n \in \mathcal{B}) \quad \mbOmega_n = \emptyset$ 
        \For{$n \in \mathcal{B}$}    
            \State \textbf{Initialize:} $\forall \, k  \quad \mbz_{nk} = 0,  \ \zeta^- = 0$
                \State $j^* \gets \argmax_{\{j\} \setminus \mbOmega_n} \ 
                \mathcal{L}_{(\cdot)}(\mbz_{\mbOmega_n} = 1, \mbz_{nj} = 1)  $
                \State $\zeta^+ = \mathcal{L}_{(\cdot)}(\mbz_{\Omega_n} = 1, \mbz_{nj*} = 1)$
                \\
                \If{$\zeta^{+} > \zeta^{-}$} 
                    \State $\mbOmega_n  \gets \mbOmega_n \cup \{ j^* \}$
                    \State $\zeta^{-} \gets \zeta^{+}$
                \Else
                    \State \Break
                \EndIf
        \EndFor
        \\
        \LeftComment{\textcolor{dimgray}{Gradient update for global parameters $\mbtheta$}}
        \State $\mbtheta \gets \textsc{adam}(\widehat{\mathcal{L}}(\mbtheta), \text{stepsize} = \rho)$ 
    \EndWhile
\end{algorithmic}
\end{algorithm}

For the Poisson likelihood model we compute the expectation:
\begin{align*}
    \mathbb{E}_{q(\mblambda)}[\ln p(\mbx,\mblambda \mid  & \mbtheta, \mbz) ] = \\ 
         &\mbx_n^{(w)} \ln \mbphi_n^{(w)} - 
         \frac{a+ \sum_w \mbx_{n}^{(w)}}{b+1}  \mbphi_n^{(w)} 
\end{align*}

To optimize $\mathcal{L}_{(\cdot)}(\mbz)$, we employ a greedy pursuit algorithm, which is similar to the matching pursuit used by K-SVD \citep{aharon2006k}. We use $\mbz_{\mbOmega_n}$ to denote a $k$-vector, corresponding to the latent vector for the $n\ts{th}$ data point, where,  $\forall j \in \mbOmega_n, \mbz_{nj} = 1$ and $\forall j \not\in \mbOmega_n,  \mbz_{nj} = 0$. To compute the sparse code given a data point $\mbx_n$, we start with an empty active set $\mbOmega_n$,  then $\forall j \in \{1,\ldots, K\}$, we individually set each $\mbz_{nj} = 1$ to find $j^* \in \{1,\ldots,K\} \setminus \mbOmega_n$ that maximizes $\mathcal{L}_{(\cdot)}(\mbz_{\mbOmega_n \cup \{j^*\}})$. We compute the scores $\zeta^{+} \triangleq \mathcal{L}_{(\cdot)}(z_{\mbOmega_n \cup \{j^*\}})$ and
$\zeta^{-} \triangleq \mathcal{L}_{(\cdot)}(z_{\mbOmega_n},\mbtheta)$. We add  $j^*$ to $\mbOmega_n$ only if $\zeta^{+} > \zeta^{-}$, this step is necessary because unlike matching pursuit, the neural net $f_\theta(\cdot)$ is a non-linear mapping from $\mbz_{\mbOmega_n}$, hence, adding  $j^*$ to $\mbOmega_n$ can decrease  $\mathcal{L}_{(\cdot)}(\mbz_{\mbOmega_n})$. For each $\mbx_n$, we repeat the preceding greedy steps to sequentially add factors to $\mbOmega_n$ till $\mathcal{L}_{(\cdot)}(\mbz_{\mbOmega_n})$ ceases to monotonically increase. 

\begin{figure*}[t]
    \centering
    \includegraphics[width=1\textwidth, trim={2.7cm 4.5cm 2.7cm 4.5cm},clip ]{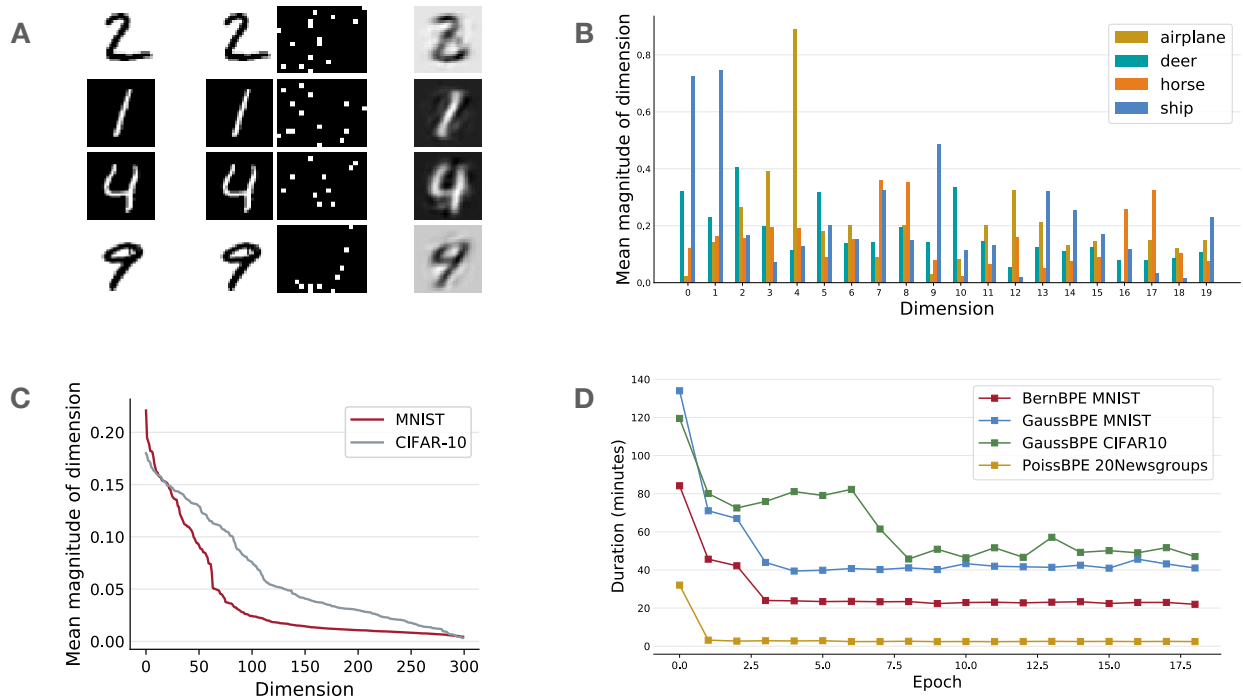}
    \caption{(A) Left: samples from scaled MNIST data. Middle: the corresponding sparse codes (reshaped as matrix for visualization), and the reconstructions using GaussBPE. Right: reconstructions using the VAE. (B) The probability of activation of the most class discriminative latent dimensions for randomly chosen classes from the CIFAR-10 dataset. As expected, the CIFAR-10 dataset utilizes more latent factors relative to the simpler MNIST dataset. (C) Sorted mean activation probabilities for all latent dimensions, for MNIST and CIFAR-10 datasets.  (D) Time duration per epoch during training for all models.}
\end{figure*}

The expected log prior on $\mbz$ imposes an approximate beta process penalty. Low probability factors learned through $q(\mbpi)$ lead to negative scores, and hence eliminate latent factors, encouraging sparse encodings $\mbz_n$. During optimization as  $q(\mbpi_k)$ for a given dimension $k$ decreases, the likelihood that the $k^{th}$ dimension will be utilized to encode the data point also decreases. Consequently, as training progresses, this allows for speed up of the sparse coding routine over iterations.

\begin{figure*}[t]
    \centering
    \includegraphics[width=.7\textwidth, trim={0.cm 6cm 0cm 6.cm},clip ]{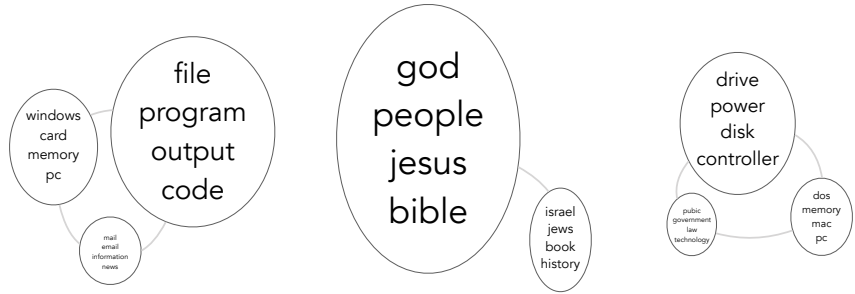}
    \caption{Learned topic associations through sparse codes. Each node represents a topic. The words within each node are the most representative words chosen from the top 15 most probable words from that topic. Groups of nodes connected by edges denote topics activated by the same sparse code, where the size of the node is proportional to the probability of the topic.}
    \label{fig:topics_asscn}
\end{figure*} 

\begin{table*}[b!]
\center
\begin{small}
\caption{Comparison of reconstruction errors and latent code sparsity on held-out data.}
\begin{tabular}{l c c l c c c c c c }
\toprule
  Dataset & \multicolumn{2}{c}{MNIST} & & \multicolumn{2}{c}{Scaled MNIST} & \multicolumn{2}{c}{ MNIST} & \multicolumn{2}{c}{CIFAR 10} \\
  Model   & GS-VAE & BernBPE & & VAE & GaussBPE & VAE & GaussBPE & VAE & GaussBPE  \\
\cmidrule(lr){2-3}\cmidrule(lr){5-10}
  NLL         &  \textbf{81.55} &82.16 & MSE      & 32.92 & \textbf{9.18} &  16.94 & \textbf{8.51} & 79.15 & \textbf{75.88}  \\
  Sparsity    & 0.40 & \textbf{0.93} & Sparsity & 0.72 & \textbf{0.86} &  0.83 & \textbf{0.86} &0.81 &  \textbf{0.96} \\
\bottomrule
\end{tabular}\label{tab:mnist1}
\end{small}
\end{table*}

\subsection{Update for $\mbtheta$}
To update the global parameters $\mbtheta$, for each model we marginalize the local scale variables $\mblambda$ and global $\mbpi$ to compute the complete data joint likelihood lower bounds $\mathcal{\widehat{L}}_\mathcal{G}, \mathcal{\widehat{L}}_\mathcal{P}, \mathcal{\widehat{L}}_\mathcal{B}$ for Gaussian, Poisson and Bernoulli likelihood models respectively, which include terms that only depend on $\mbtheta$:
\begin{align*}
    \mathcal{\widehat{L}_\mathcal{G}}(\mbtheta) &= 
        \mathbb{E}_{q(\mblambda)}\left[\ln p(\mbx,\mblambda \mid  \mbtheta, \mbz) \right] \\
    \mathcal{\widehat{L}_\mathcal{P}}(\mbtheta) &= 
        \mathbb{E}_{q(\mblambda)}\left[\ln p(\mbx,\mblambda \mid  \mbtheta, \mbz) \right] \\
    \mathcal{\widehat{L}_\mathcal{B}}(\mbtheta) &= 
        \ln p(\mbx \mid \mbtheta, \mbz) 
\end{align*}
For Gaussian:
\begin{align*}
    \mathbb{E}_{q(\mblambda)} \left[ \ln p(\mbx,  \mblambda \mid \mbtheta, \mbz) \right] =  \| \mbx_n -  \mu_{\mblambda_n \mid \mbx_n, \mbz_n, \mbtheta} \cdot f_\theta(\mbz_n) \|^2 \\
    + \sigma^{2}_{\mblambda_n \mid \mbx_n, \mbz_n, \mbtheta} \cdot f_\theta(\mbz_n)^\top f_\theta(\mbz_n) + c
\end{align*}

For Poisson and Bernoulli the likelihood is same as that in sparse coding step. We use stochastic optimization to update $\mbtheta$ using ADAM \cite{kingma2014adam}. First order gradient methods with moment estimates such as ADAM, can implicitly take into account the rate of change of natural parameters $(\mba_k , \mbb_k)$ for $q(\mbpi)$ when optimizing the neural net parameters. The full sparse coding algorithm is outlined in Algorithm \ref{alg:sparse_code}.

\section{Empirical study}
We demonstrate the potential of our beta process sparse encodoing models in a variety of settings. We evaluate the Gaussial likelihood Beta-Bernoulli Process Encoder (GaussBPE) on scaled MNIST \citep{lecun2010mnist} and CIFAR-10 \citep{Krizhevsky09learningmultiple} datasets. The scaled MNIST dataset consists of MNIST images that are randomly scaled using a scaling factor sampled from $\mathcal{U}(-\text{scale max}, \text{scale max})$. We evaluate the BernBPE on MNIST data.  To compare GaussBPE to Gaussian VAE and BernBPE to Gumbel-Softmax VAE, we compare the sparsity of the learned encodings, as well as the reconstruction error on held-out data. We utilize the following metrics:

Lastly, we present qualitative results for the PoissonBPE on 20-Newsgroup dataset \citep{joachims1996probabilistic} to uncover latent distributions over topics.

\subsection{Sparsity}
 We quantify the sparsity of the inferred latent encodings using the Hoyer extrinsic metric    ~\citep{hurley2009comparing}, which is $0$ for a fully dense vector and $1$ for a fully sparse vector. For a set of latent encodings $\{\mbz_n\}_{n=1}^{N}$, the sparsity is defined as:
\begin{align*}
\mathrm{Sparsity}(\{\mbz_n\}_{n=1}^{N}) = \frac{1}{N} \sum_n \mathrm{Hoyer}(\mbz_n) \\
\mathrm{Hoyer}(\mbz_n) = \frac{\sqrt{K} - \| \mbz_n \|_1 / \| \mbz_n \|_2}{\sqrt{K} - 1} \in [0, 1] 
\end{align*}
For the VAE models, we use the encoding means $\{\mathbb{E}_{q(\mbz_n | \mbx_n)}[\mbz_n]\}_{n=1}^{N}$ in lieu of $\{\mbz_n\}_{n=1}^{N}$.
\subsection{Reconstruction Error}
For GaussBPE and Gaussian likelihood VAE, we report the reconstruction mean squared error (MSE). For the Gaussian likelihood VAE, $\mathbb{E}[\mblambda_n] = 1$, and we use $\mathbb{E}_{q(\mbz_n | \mbx_n)}[f_\theta(\mbz_n)]$ instead of $f_\theta(\mbz_n)$:
\begin{align*}
\mathrm{MSE}(\{\mbx_n, \mbz_n\}_{n=1}^{N}) = \frac{1}{N} \| \mbx_n - \mathbb{E}[\mblambda_n]f_\theta(\mbz_n) \|^2 
\end{align*}

For the BernBPE and Bernoulli likelihood Gumbel Softmax VAE, we report the negative log likelihood (NLL):
\begin{align*}
    \mathrm{NLL}(\{\mbx_n, \mbz_n\}_{n=1}^{N}) = - \frac{1}{N} \ln  p(\mbx_n | \mbz_n)
\end{align*}

For the VAE models, we use the same recognition network architecture as the original papers. For the VAE likelihood models and the GaussBPE and BernBPE likelihood models, we use the same architecture as that used by the Gumbel Softmax VAE paper. Notably, the last layer is linear for Gaussian VAE, however, sigmoid for GaussBPE, as in our model, $\mblambda_n$ decouples the scaling of individual data points. A summary of all the hyperparameters used for all models can be found in the supplementary material.

We evaluate the PoissBPE model on 20-Newsgroup data. We pre-process the data by removing headers, footers and quotes, as well as English stop words to get a $512$ dimensional vocabulary. We then vectorize each document to a $1142$ dimensional vector, where each dimension represents the number of occurrences of a particular word in the vocabulary, within the document. For the PoissBPE, we choose a $W \times T$ $\mbbeta$ matrix, with $W=1142$ and $T=15$, this corresponds to a topic model with $15$ topics, where each topic vector $\mbbeta_{:,t}$, is a distribution over the $1142$ words. The last layer non-linearity is a softmax, hence, the $f_\theta(\mbz_n)$, maps $\mbz_n$ to a probability distribution over the $T$ topics. 

\section{Results}
On binary MNIST data, where the scale of the data points does not affect the latent encodings, we found the BernBPE model to be comparable to the Gumbel Softmax VAE in terms of reconstruction error, however, it does so by utilizing substantially fewer latent dimensions. For real valued MNIST data, the GaussBPE significantly outperformed the Gaussian likelihood VAE in terms of both the reconstruction error as well as the sparsity of the latent codes. For randomly scaled MNIST data, the relative improvement in sparsity was similar to the improvement observed over VAE on real valued MNIST data, however, the reconstruction error was markedly better. Lastly, on the CIFAR-10 dataset, the GaussBPE performed better than VAE in terms of reconstruction error and sparsity. We summarize our experimental results in Table \ref{tab:mnist1}.

\section{Discussion}
We evaluated our models across four datasets to explore the effects of the different variables we introduce in our generative model. On the binary MNIST data, where scale is not a factor, as expected, we observed similar performance in terms of reconstruction error, however, the Beta-Bernoulli process prior encouraged spasrity in the latent representation, which lead to our model to learn much sparser representaions. For real valued MNIST data, with variations in intensity across images, the local scaling variable allowed the model to learn sparser encodings, while also improving the reconstruction error. We further explored this effect by exaggerating the local scale variations by randomly perturbing the intensity of the MNIST images. As we expected, this lead to significant deterioration in image reconstructions by the VAE. Our explicit modeling of local variations decoupled the local data scaling from the encoding process, which allowed the model to learn scale invariant encodings, resulting in substantially improved performance over the VAE. On natural image datasets such as CIFAR-10, we expect more variation in image intensity relative to the standard MNIST dataset. Since our model performs well even under random perturbation local data scaling, we expected the GaussBPE to perform well on the CIFAR-10 dataset. As we had hoped, the our model learned sparser encodings while also improving reconstruction error on the CIFAR-10 dataset.

\bibliography{main}
\bibliographystyle{abbrvnat}
\end{document}